\documentclass{article}
\usepackage{gaussifier,times}
\usepackage[T1]{fontenc}
\usepackage{amsmath,amssymb,bm,amsthm}
\usepackage{graphicx,booktabs}
\usepackage{microtype}
\usepackage{hyperref}
\usepackage{xurl}
\usepackage{enumitem}
\hypersetup{hidelinks}
\usepackage{multirow}
\usepackage{algorithm}
\usepackage{algpseudocode}

\title{Gaussifier: Fast 2D Gaussian Decomposition}

\author{%
\textbf{Yunlai Zhou}$^{1,2}$ \quad
\textbf{Yu Yin}$^{2}$ \quad
\textbf{Michael W. Jenkins}$^{1,2}$ \quad
\textbf{Yehe Liu}$^{1,2}$ \\[0.4em]
{\normalfont
$^{1}$OpsiClear LLC \qquad
$^{2}$Case Western Reserve University
} \\[0.3em]
{\normalfont\small
\texttt{\{yunlai,michael,yehe\}@opsiclear.com} \quad
\texttt{yu.yin@case.edu}
}
}

\newcommand{\method}{Gaussifier}
\newcommand{\emv}{EMV}
\newcommand{\cL}{\mathcal{L}}
\newcommand{\clip}{\operatorname{clip}}
\newcommand{\sampleat}{\mathcal{B}}

\begin{document}
\maketitle
\lhead{Preprint}
\begin{abstract}
2D Gaussians render images at thousands of frames per second.
However, fitting an image takes seconds of optimization.
This confines 2D Gaussians to offline use.
Feedforward methods are faster but less accurate without per-image fine-tuning.
We present Gaussifier, a lightweight feedforward framework for 2D Gaussian decomposition.
We learn a density map of Gaussian centers from per-image fitting data.
Equal-mass Voronoi places Gaussians so that each Voronoi cell holds approximately equal density.
A stateless, reusable network then refines the output in a few iterations.
Gaussifier reaches 40.73/38.61\,dB PSNR on Kodak-24/DIV2K at over 60 FPS.
At the same Gaussian count, it exceeds every feedforward baseline by 3.0 to 15.6\,dB in PSNR.
With fewer Gaussians than any baseline, it still leads the strongest one by 2.5 to 3.5\,dB.
At the same count, baselines with their own per-image fitting need 26 to 326 times longer to reach its PSNR.
Gaussifier makes high-fidelity 2D Gaussian images fast and light enough for real-time applications.
We release code, weights and a browser-based WebGPU demo at \url{https://github.com/Gaussifier/gaussifier}.
\end{abstract}

\section{Introduction}
\label{sec:introduction}
Gaussian images represent an image as a set of explicit 2D Gaussian primitives, each with its own position, shape, and color~\citep{zhang2024gaussianimage,zhang2025imagegs,li2026gaussianimagepp}.
These sparse representations are increasingly used in many applications, including vision-language models~\citep{omri2026gaussianvision}, image restoration~\citep{chen2026llgaussianimage}, and video coding~\citep{wang2025gsvc}.
However, fitting Gaussians to each image takes seconds of optimization.
Existing feedforward models rely on per-image fine-tuning or give up quality~\citep{zeng2025instantgi,wang2026fast2dgs,zeng2026air}.
So far, high-fidelity 2D Gaussian representations remain limited to offline use because of this bottleneck.

An ideal decomposition reconstructs the input faithfully with few primitives and runs fast.
Existing methods trade these goals against one another.
For example, a codec trades quality for size.
Furthermore, decomposing an image into Gaussians is an ill-posed problem.
Many different sets of Gaussians can render nearly the same image.
The decomposition must jointly decide the Gaussian count, locations, shapes, and colors.
Since overlapping Gaussians can contribute to the same pixel, the best parameters for one Gaussian depend on its neighbors.
Fitting resolves these dependencies with thousands of gradient steps. Feedforward methods still rely on this fitting for final polishing.

We present Gaussifier, a lightweight feedforward framework.
It learns a density map of Gaussian centers from per-image fits, places Gaussians on it, and refines them with a few learned updates.
At the same count, it exceeds every feedforward baseline in peak signal-to-noise ratio (PSNR) by 3.0 to 15.6\,dB, and baselines need orders of magnitude more fitting time to match its PSNR.
At its predicted count, it leads every baseline by over 8\,dB.
We make three contributions:

\begin{itemize}[leftmargin=*,itemsep=1pt,parsep=0pt]
    \item \textbf{Learned sparse allocation.} A lightweight model predicts the expected number of Gaussian centers at each pixel. The sum of this density map sets the Gaussian count, which users can scale up or down. Equal-mass Voronoi (EMV) sampling places the Gaussians so that each Voronoi cell holds approximately equal density mass. The resulting layout resembles that of the fitted Gaussians.
    \item \textbf{Stateless refinement without fitting gradients.} One refinement network jointly updates all Gaussians from the rendering residual. Six updates add about 15 to 16\,dB. The same weights serve any Gaussian count, and one to six updates trade speed for quality.
    \item \textbf{Video-rate and high-fidelity decomposition.} Gaussifier decomposes a $512\times512$ image in under 15\,ms on average and reaches 40.73/38.61\,dB PSNR on Kodak-24/DIV2K. On Kodak-24, at the same Gaussian count, it matches the PSNR achieved by GaussianImage with 20\,s of fitting. This moves 2D Gaussian images from offline to real-time and large-scale use.
\end{itemize}

\section{Related Work}
\label{sec:related}

\paragraph{Optimization-based Gaussian image representations.}
Implicit neural representations fit a network to each image and evaluate it at every pixel~\citep{sitzmann2020siren,muller2022instantngp,dupont2021coin}.
GaussianImage~\citep{zhang2024gaussianimage} brings 3D Gaussian Splatting (3DGS)~\citep{kerbl2023gaussians} to images and renders a fitted image at over 1,000 frames per second (FPS).
Image-GS~\citep{zhang2025imagegs}, GaussianImage++~\citep{li2026gaussianimagepp}, and structure-guided allocation~\citep{liang2025structure} improve the placement of the Gaussians with error-guided growth, densification, and structural priors.
All of these methods fit each image by iterative optimization, which takes seconds per image.

\paragraph{Feedforward Gaussian images.}
Feedforward approaches predict Gaussian parameters with a trained network instead of per-image optimization.
Instant-GI dithers a predicted probability map of Gaussian positions into points~\citep{zeng2025instantgi}.
Capacity-constrained point distributions instead give every point an equal share of the density~\citep{balzer2009capacity,degoes2012bnot}.
Fast-2DGS samples Gaussian positions from a spatial prior conditioned on the Gaussian count. A separate network then predicts dense Gaussian attributes~\citep{wang2026fast2dgs}.
Instant-GI and Fast-2DGS both need per-image fine-tuning to reach high fidelity.
AIR avoids optimization by adding Gaussians stage by stage from the reconstruction residual~\citep{zeng2026air}.
However, earlier Gaussians stay fixed, so later stages cannot revise them.
Feedforward 3DGS predicts Gaussians anchored to the pixels of one or a few input images~\citep{szymanowicz2024splatter,charatan2024pixelsplat,chen2024mvsplat,zhang2024gslrm}.
ReSplat and GIFSplat can refine these Gaussians iteratively from rendering errors without test-time gradients~\citep{xu2026resplat,chen2026gifsplat}.

\paragraph{Gaussian image applications.}
Gaussian primitives are compact, editable, and fast to render.
Gaussian tokenizers~\citep{dong2025gaussiantoken,zhang2025gpstoken} and super-resolution networks~\citep{hu2024gaussiansr,chen2025gsasr} predict Gaussians inside a trained model.
Other methods decompose the input image.
GViT classifies images from a few hundred Gaussians optimized jointly with the classifier~\citep{hernandez2025gvit}.
GaussianVision aligns Gaussian images with language after fitting 12.8M images~\citep{omri2026gaussianvision}.
Zero-shot restoration fits a low-light or hazy image for 20-30k iterations and adjusts its Gaussians~\citep{chen2026llgaussianimage,chen2026llgaussianmap,chen2026dehazegaussianimage}.
MiraGe edits images through flat Gaussians fitted to the image~\citep{waczynska2024mirage}.
Gaussian codecs transmit quantized fitted Gaussians~\citep{zhang2024gaussianimage,wang2025gsvc}.
GSVC decodes at over 1,500 FPS but takes about 200\,s to encode each 1080p frame~\citep{wang2025gsvc}.
In these decomposition methods, producing the Gaussians is a major bottleneck because each image or frame needs its own optimization.

\section{Method}
\label{sec:method}

\begin{figure}[!htbp]
    \centering
    \includegraphics[width=\linewidth]{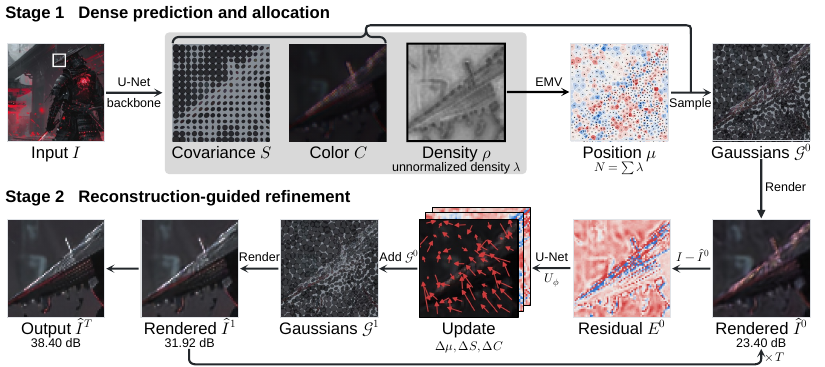}
    \caption{\textbf{Overview of \method{}.} Panels after the input show the region in its white box. In Stage 1, a U-Net maps the image $I$ to log-covariance, color, and density maps $S$, $C$, and $\lambda$. The sum of $\lambda$ gives the Gaussian count $N$. \emv{} places $N$ centers $\boldsymbol\mu$ from the normalized density $\rho$. Sampling $S$ and $C$ at these centers gives the initial Gaussians $\mathcal{G}^0$. In Stage 2, a stateless refinement network $U_\phi$ reads the image, the Stage-1 maps, the current rendering $\widehat I^t$, its residual $E^t$, and the count map $H^t$. For simplicity, the diagram shows only the residual input. It predicts updates $\Delta\boldsymbol\mu$, $\Delta\boldsymbol S$, and $\Delta\boldsymbol C$ to the center, log-covariance, and color of every Gaussian, while $N$ stays fixed. This repeats $T$ times. Displacement arrows in $\Delta\boldsymbol\mu$ are magnified for visibility.}
    \label{fig:method_overview}
\end{figure}

\subsection{Gaussian Representation and Method Overview}
\label{subsec:gaussian_representation_and_method_overview}
Given an image $I$ on a pixel grid $\Omega$ with $P$ pixels, we can construct $N$ anisotropic 2D Gaussians.
Each Gaussian has a continuous position $\boldsymbol\mu_i$, a covariance $\boldsymbol\Sigma_i$, and a color coefficient $\boldsymbol a_i$.
Following additive Gaussian image representations~\citep{zhang2024gaussianimage}, we reconstruct the image as
\begin{equation}
    \widehat I(\boldsymbol x)
    = \clip_{[0,1]}\!\left[
        \sum_{i=1}^{N}\boldsymbol a_i
        \exp\!\left(-\frac{1}{2}(\boldsymbol x-\boldsymbol\mu_i)^\top
        \boldsymbol\Sigma_i^{-1}(\boldsymbol x-\boldsymbol\mu_i)\right)
    \right].
    \label{eq:gaussian_render}
\end{equation}
We parameterize the covariance by its matrix logarithm $\boldsymbol S_i=\log\boldsymbol\Sigma_i$~\citep{arsigny2006log}.
The covariance $\boldsymbol\Sigma_i=\operatorname{expm}(\boldsymbol S_i)$ then stays positive definite under additive updates.
For rendering, we convert $\boldsymbol S_i$ to the equivalent principal scale and rotation. Based on this representation, our framework is illustrated in Figure~\ref{fig:method_overview}.

\subsection{Gaussian Sampling}
\label{subsec:gaussian_sampling}
\paragraph{Design.}
Gradient-based optimization produces well-placed Gaussians in both 2D and 3D splatting~\citep{zhang2024gaussianimage,kerbl2023gaussians}.
Feedforward models trained only with a photometric loss often place Gaussians in repeating patterns that do not follow the image content.
This degrades their visual quality.
We therefore use Gaussians fitted by per-image optimization as teachers for learned sampling.
We first predict a Gaussian density map from the input image.
A sampler then places Gaussians according to this map.
Their layout resembles that of the optimized 2D Gaussians.

\paragraph{Density map.}
For efficiency, a 2M-parameter U-Net~\citep{ronneberger2015unet} predicts the expected number of Gaussian centers in each pixel, $\lambda(\boldsymbol x)$.
It also predicts the initial log-covariance $S(\boldsymbol x)$ and color $C(\boldsymbol x)$ of a Gaussian centered at each pixel $\boldsymbol x$.
Inspired by the training strategy in bit2bit~\citep{liu2024bit2bit}, we train $\lambda$ using sparse rasterized targets that equal 1 at the centers of fitted reference Gaussians and 0 elsewhere.
The exact positions of fitted centers vary even between similar image regions.
The prediction $\lambda$ therefore becomes their expected density, which is smooth.
This density carries two pieces of information.
Its intensity profile tells where the Gaussians should be.
Its sum defines a total Gaussian count, $N=\sum_{\boldsymbol x\in\Omega}\lambda(\boldsymbol x)$, as in density-based object counting~\citep{lempitsky2010learning}.
We keep only the intensity profile by dividing $\lambda$ by its mean, $\rho(\boldsymbol x)=P\,\lambda(\boldsymbol x)/N$.
The relative density $\rho$ averages to one, so $\rho(\boldsymbol x)=2$ marks a region that needs twice the average density of Gaussians.
A user can also scale $N$ up or down while keeping the same $\rho$ (Appendix~\ref{app:fixed_budgets}).

\paragraph{Equal-mass Voronoi sampling.}
EMV turns $\rho$ into $N$ Gaussian centers by treating placement as a coverage problem.
We partition the image into $N$ Voronoi cells $V_i$ with one center in each.
Every cell should hold approximately the same density mass, $m_i=\sum_{\boldsymbol x\in V_i}\rho(\boldsymbol x)\approx\bar m$, where $\bar m$ is the mean cell mass.
Highly textured regions receive both more and smaller cells.
This is similar to capacity-constrained Voronoi tessellations, which enforce equal mass exactly but require many iterations~\citep{balzer2009capacity}.
Random sampling leaves clumps and gaps, while dithering matches the local count but the cells around its centers still hold unequal mass.
Lloyd relaxation alone spreads centers in proportion to $\rho^{1/2}$ and under-serves detailed regions~\citep{lloyd1982quantization,du1999cvt}.
Inspired by weighted Linde-Buzo-Gray stippling~\citep{deussen2017wlbg}, EMV approximates equal mass by oversampling sites from $\rho$ and merging the lightest cells (Algorithm~\ref{alg:emv_placement} and Appendix~\ref{app:emv_details}).
Weighted Linde-Buzo-Gray stippling splits heavy cells and deletes light ones over many iterations.
EMV never splits cells.
It starts from about $1.5N$ sites and merges light cells into their neighbors until $N$ remain.
A few Lloyd-relaxation iterations then even out the local spacing and cell masses.
The sites become the initial centers $\boldsymbol\mu_i^0$, and the final cells give the masses $m_i$.
Placement takes about 2 to 3\,ms for 40k-70k Gaussians on an RTX 5090.

\begin{algorithm}[t]
\caption{Approximate \emv{} placement in pixel coordinates.}
\label{alg:emv_placement}
\begin{algorithmic}[1]
\Require Density $\rho$ on $\Omega$ and Gaussian count $N$
\Ensure Initial centers $\{\boldsymbol\mu_i^0\}$ and cells $\{V_i\}$
\State Sample about $1.5N$ sites $Q=\{\boldsymbol q_i\}$ from $\rho$ with stratified sampling
\While{$|Q|>N$}
    \State Compute the Voronoi cells $\{V_i\}$ of $Q$ and their masses $m_i=\sum_{\boldsymbol x\in V_i}\rho(\boldsymbol x)$
    \State Merge the lowest-mass cells into their lowest-mass neighbors
    \State Move each kept site to the mass-weighted mean of itself and the sites merged into it
\EndWhile
\For{$k=1,2,3$}
    \State Recompute $\{V_i\}$ and move each nonempty site to $\sum_{\boldsymbol x\in V_i}\rho(\boldsymbol x)\boldsymbol x\big/\sum_{\boldsymbol x\in V_i}\rho(\boldsymbol x)$
\EndFor
\State Recompute $\{V_i\}$ for the final sites
\State \Return $\boldsymbol\mu_i^0=\boldsymbol q_i$ and the cells $\{V_i\}$
\end{algorithmic}
\end{algorithm}

\paragraph{Initial Gaussians.}
Each Gaussian takes its initial log-covariance and color from $S$ and $C$ at its center by bilinear interpolation $\sampleat$, $\boldsymbol S_i^0=\sampleat(S,\boldsymbol\mu_i^0)$ and $\boldsymbol C_i^0=\sampleat(C,\boldsymbol\mu_i^0)$.
Because EMV chooses the centers by discrete sampling and merging, reconstruction gradients cannot reach the density map through the center locations.
We therefore scale the color of each Gaussian by its relative cell mass, $w_i=m_i/\bar m$.
Each $m_i$ sums $\rho$ over its cell, so the reconstruction gradient on $w_i$ reaches every pixel of the cell.
This gradient adjusts the shape of the density but not the count (Appendix~\ref{app:emv_details}).
During pretraining, a density loss (Eq.~\ref{eq:density_loss}) also supervises $\lambda$ directly.
The cell-mass weights are computed once and stay fixed while the centers move.
However, the refinement network described in the next section still updates the unweighted color $\boldsymbol C_i^t$, while Gaussian $i$ uses the color coefficient $\boldsymbol a_i=\clip_{[0,1]}(w_i\boldsymbol C_i^t)$ in Eq.~\ref{eq:gaussian_render} to render images.

\subsection{Reconstruction-Guided Refinement}
\label{subsec:reconstruction_guided_recurrent_refinement}
\paragraph{Design.}
EMV gives a good initial approximate allocation.
However, overlapping Gaussians contribute to the same pixel, so the attributes of each Gaussian depend on its neighbors.
Per-image fitting resolves this with thousands of gradient steps.
Each step is driven by the rendering loss and is slow.
To avoid this, we train a network that reads the errors and updates all Gaussians together, iteratively.
The network can learn to coordinate attribute changes from millions of images, so that a few inference iterations can replace about a hundred gradient steps.

\paragraph{Update network.}
At step $t$, we render the current Gaussians to obtain $\widehat I^t$ and the residual $E^t=I-\widehat I^t$.
We also rasterize the current centers into a count map $H^t$, which shows where the Gaussians currently lie.
The network $U_\phi$ reads these maps and the initialization context $Z=[I,\rho,S,C]$.
It predicts a dense update field, which we sample at each center to update the Gaussians:
\begin{align}
    D^t&=U_\phi\!\left([Z,\widehat I^t,E^t,H^t]\right),\qquad
    (\Delta\boldsymbol\mu_i^t,\Delta\boldsymbol S_i^t,\Delta\boldsymbol C_i^t)
    =\sampleat(D^t,\boldsymbol\mu_i^t),
    \label{eq:correction_field}\\
    \boldsymbol\mu_i^{t+1}&=\Pi_{\mathrm{img}}(\boldsymbol\mu_i^t+\Delta\boldsymbol\mu_i^t),\qquad
    \boldsymbol S_i^{t+1}=\boldsymbol S_i^t+\Delta\boldsymbol S_i^t,\qquad
    \boldsymbol C_i^{t+1}=\clip_{[0,1]}(\boldsymbol C_i^t+\Delta\boldsymbol C_i^t),
    \label{eq:parameter_updates}
\end{align}
where $\Pi_{\mathrm{img}}$ projects centers onto the image domain and $\Delta\boldsymbol S_i^t$ is symmetric.

\paragraph{Stateless updates.}
Our refinement is inspired by ReSplat, which refines 3D Gaussians from rendering errors~\citep{xu2026resplat}.
ReSplat keeps a hidden state for every Gaussian and updates it with a transformer that attends over neighbors.
Its cost grows with the number of Gaussians.
Our network is stateless and carries no hidden state between updates.
We use a small U-Net with 0.75M parameters instead of a transformer.
It reads and predicts image-sized maps, so the same set of weights works for any number of Gaussians (Appendix~\ref{app:fixed_budgets}), and its cost does not grow with the Gaussian count.
All $T=6$ updates share the same parameters $\phi$, and each update costs one pass of the refinement network and one rendering.

\subsection{Training Objectives}
\label{subsec:training_objectives_and_schedule}
\paragraph{Pretraining data.}
We pretrain the 2M-parameter U-Net on Open Images data~\citep{kuznetsova2020openimages}.
Each image is paired with reference Gaussians fitted to it by per-image optimization.
We first train on about 107k such pairs.
We then fine-tune on about 20k pairs whose fits start from an earlier Gaussifier prediction (Appendix~\ref{app:reference_fits}).  %
Each of the three output maps has its own target.

\paragraph{Density.}
The density target $y$ is a sparse map with a unit mass at each reference Gaussian center and zero elsewhere.
We splat each center bilinearly onto its four nearest pixels to keep sub-pixel positions.
The sum of $y$ therefore equals the reference count.
We train $\lambda$ with a Poisson loss,
\begin{equation}
    \cL_{\mathrm{density}}
    =\frac{1}{P}\sum_{\boldsymbol x\in\Omega}
    \left[\lambda(\boldsymbol x)-y(\boldsymbol x)\log\lambda(\boldsymbol x)\right].
    \label{eq:density_loss}
\end{equation}
Eq.~\ref{eq:density_loss} supervises only the local rate of Gaussian centers and count but does not constrain their spacing, which EMV then sets (Section~\ref{subsec:gaussian_sampling}).

\paragraph{Log-covariance and color.}
For each reference Gaussian, we build its covariance from its fitted scale and rotation and take the matrix logarithm.
We splat the three entries onto the same four pixels as its center and average them at each pixel with the bilinear weights.
This gives the dense target $\bar S$.
The color target $\bar C$ is built the same way from the fitted color coefficients.
We do not use pixel colors, because additive rendering makes each pixel a sum of overlapping Gaussians.
Both maps use a smooth $L_1$ loss $\ell$ averaged with the weights $y$, so only pixels near reference centers contribute,
\begin{equation}
    \cL_S=\frac{\sum_{\boldsymbol x}y(\boldsymbol x)\,\ell\big(S(\boldsymbol x)-\bar S(\boldsymbol x)\big)}{\sum_{\boldsymbol x}y(\boldsymbol x)},
    \label{eq:attribute_loss}
\end{equation}
and $\cL_C$ is defined in the same way.
The pretraining loss is $\cL_{\mathrm{density}}+0.05\,\cL_S+0.1\,\cL_C$.

\paragraph{Reconstruction.}
Next, we train the refinement network using only images.
Our objective supervises the initial reconstruction and the reconstruction after each update,
\begin{equation}
    \cL_{\mathrm{rec}}
    =\sum_{t=0}^{T}\alpha_t\operatorname{MSE}(\widehat I^t,I),
    \qquad
    \alpha_t=\frac{t+1}{\sum_{s=0}^{T}(s+1)}.
    \label{eq:reconstruction_loss}
\end{equation}
The increasing weights emphasize the later reconstructions.
This stage uses about 3.55M training images in Open Images with a fixed 99:1 training and validation split.
We also fine-tune the output heads of the pretrained U-Net and keep its other layers frozen (Appendix~\ref{app:final_adaptation}).

\section{Experiments}
\label{sec:experiments}

\subsection{Experimental Setup}
\label{subsec:experimental_setup}

\paragraph{Datasets.}
We evaluate on Kodak-24~\citep{franzen1999kodak}, which contains 24 images, and on the 100 high-resolution images in the DIV2K validation set~\citep{agustsson2017div2k}.
Both datasets use the same preprocessing.
We resize the shorter image side to 512 pixels with bilinear interpolation and take a $512\times512$ center crop.

\paragraph{Baselines and evaluation metrics.}
We consider two groups of baselines.
For feedforward reconstruction, we compare with AIR~\citep{zeng2026air}, Instant-GI~\citep{zeng2025instantgi}, and Fast-2DGS~\citep{wang2026fast2dgs}.
We use their released implementations and checkpoints.
We also evaluate Instant-GI$^*$ and Fast-2DGS$^*$, obtained by continuing training from their official checkpoints on the same Open Images training images as Gaussifier.
For optimization-based methods, we compare with GaussianImage~\citep{zhang2024gaussianimage} at a matched Gaussian count (Appendix~\ref{app:gaussianimage_matched}).
We report PSNR and multi-scale SSIM (MS-SSIM)~\citep{wang2003msssim} averaged over all images, with LPIPS~\citep{zhang2018lpips} in Appendix~\ref{app:lpips}.
We measure speed and peak device memory on an NVIDIA RTX 5090.
Gaussifier runs in its TensorRT/AOTInductor deployment, and the baselines run in FP32.
Appendix~\ref{app:evaluation_details} also reports compiled baselines.

\subsection{Feedforward Reconstruction}
\label{subsec:feedforward_reconstruction}

\begin{table}[!htbp]
    \centering
    \caption{Feedforward reconstruction on Kodak-24 and DIV2K. Best and second-best results are shown in \textbf{bold} and \underline{underlined}. Mem.\ is the peak device memory in MB, and \#G is the mean Gaussian count in thousands. Higher is better except for Mem., and \#G is not ranked. $^{*}$ denotes methods fine-tuned on our training set. Gaussifier uses an image-dependent Gaussian count, and Gaussifier-20k uses a fixed low budget of 20k Gaussians.}
    \label{tab:main_table}
    \setlength{\tabcolsep}{2.6pt}
    \begin{tabular}{lccccccccc}
        \toprule
        & \multicolumn{4}{c}{Kodak-24} && \multicolumn{4}{c}{DIV2K} \\
        \cmidrule(lr){2-5}\cmidrule(lr){7-10}
        Method
        & PSNR & MS-SSIM & Mem. & \#G (k)
        && PSNR & MS-SSIM & Mem. & \#G (k) \\
        \midrule
        Gaussifier
        & \textbf{40.73} & \textbf{0.9963} & 1,126 & 42.6
        && \textbf{38.61} & \textbf{0.9973} & 1,130 & 72.2 \\
        Gaussifier-20k
        & \underline{35.97} & \underline{0.9860} & \underline{1,114} & 20.0
        && \underline{30.15} & 0.9735 & \underline{1,114} & 20.0 \\
        AIR
        & 31.86 & 0.9853 & 2,852 & 35.4
        && 27.57 & \underline{0.9810} & 2,852 & 36.3 \\
        Instant-GI
        & 30.40 & 0.9674 & 3,364 & 25.1
        && 27.41 & 0.9714 & 3,370 & 28.8 \\
        Instant-GI$^{*}$
        & 32.47 & 0.9802 & 3,364 & 28.8
        && 27.66 & 0.9799 & 3,370 & 29.0 \\
        Fast-2DGS
        & 25.95 & 0.9203 & \textbf{1,022} & 50.0
        && 23.73 & 0.9233 & \textbf{1,022} & 70.0 \\
        Fast-2DGS$^{*}$
        & 29.03 & 0.9734 & \textbf{1,022} & 50.0
        && 26.98 & 0.9607 & \textbf{1,022} & 70.0 \\
        \bottomrule
    \end{tabular}
\end{table}

\begin{figure}[!htbp]
    \centering
    \includegraphics[width=\linewidth]{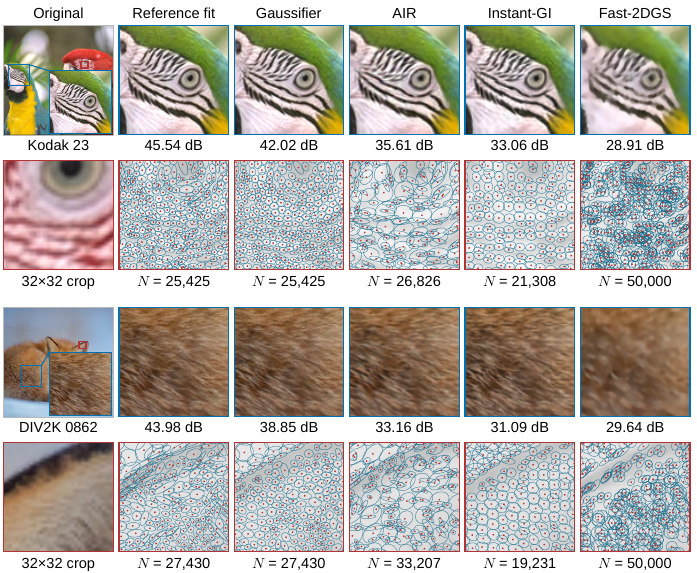}
    \caption{\textbf{Feedforward reconstruction quality and Gaussian geometry.} For each example, the first row shows the original image and magnified $96\times96$ reconstruction crops marked in blue. The crops show a per-image fit for 50k iterations as a reference and the reconstructions of Gaussifier and three feedforward baselines. The second row shows a separate $32\times32$ crop marked in red, with Gaussian centers (red) and $1\sigma$ covariance ellipses (blue). PSNR and the Gaussian count $N$ refer to the full image. Compared with the three baselines, Gaussifier produces more reasonable Gaussian geometry.}
    \label{fig:feedforward_reconstruction_quality_and_Gaussian_geometry}
\end{figure}

Table~\ref{tab:main_table} reports forward-only reconstruction quality and efficiency.
Gaussifier achieves the highest PSNR and MS-SSIM on both datasets.
With its default adaptive Gaussian count, Gaussifier reaches 40.73/38.61\,dB on Kodak-24/DIV2K, exceeding the strongest baseline, Instant-GI$^{*}$, by 8.26/10.95\,dB.
Against every released baseline, it has higher PSNR on every image, and paired 95\% bootstrap intervals exclude zero (Appendix~\ref{app:paired_stats}).
This configuration runs in real time at 73/68 FPS and uses 1.1\,GB of device memory.
Figure~\ref{fig:feedforward_reconstruction_quality_and_Gaussian_geometry} presents the qualitative comparison.

\paragraph{Gaussian budget.}
The Gaussian count of Gaussifier adapts to the image content, from 22k to 85k Gaussians on Kodak-24 and from 19k to 170k on DIV2K.
With a set budget on Kodak-24, PSNR rises from 35.97\,dB at 20k to 41.56\,dB at 50k.
At 50k, this matches GaussianImage's 41.37\,dB after 20\,s of fitting at the same Gaussian count (Appendix~\ref{app:gaussianimage_matched}).
With 20k Gaussians, fewer than any baseline in Table~\ref{tab:main_table} uses, Gaussifier still exceeds the strongest baseline by 3.50/2.49\,dB on Kodak-24/DIV2K.

\paragraph{Controlled comparisons.}
Gaussifier supports a configurable Gaussian count $N$ and number of refinement updates.
We compare it with each baseline at the same Gaussian count.
At the per-image count of each baseline, Gaussifier improves mean PSNR by 3.0 to 15.6\,dB, with higher PSNR on every image (Tables~\ref{tab:same_count} and~\ref{tab:matched}).
Additionally, at the same Gaussian count, one or two refinement updates are enough to reach or exceed the mean PSNR of AIR and Instant-GI.
Table~\ref{tab:same_count} also compares Gaussifier with baselines that perform per-image fitting at the same counts.
Gaussifier reaches its mean PSNR with six refinement updates in at most 14\,ms.
To reach the same PSNR, AIR and Instant-GI need 0.36 to 1.21\,s, and Fast-2DGS needs 4.26 to 4.56\,s.
This is 26 to 326 times longer than Gaussifier.

\begin{figure}[!htbp]
    \centering
    \includegraphics[width=\linewidth]{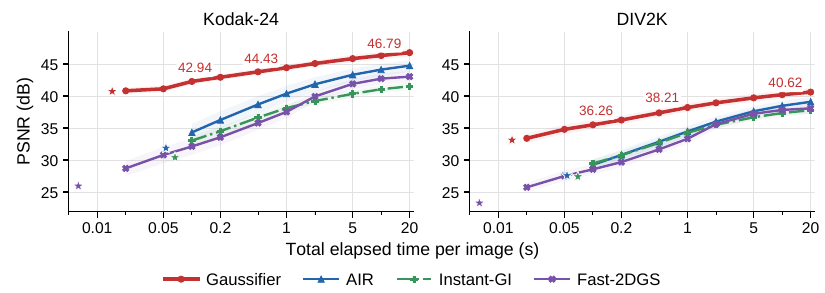}
    \caption{\textbf{Quality over time with continued per-image fitting.} On Kodak-24, each method fits at its default Gaussian count (Table~\ref{tab:main_table}). On DIV2K, Gaussifier fits at the lowest per-image count among the three baselines. Fast-2DGS uses 50k on both datasets. Stars mark feedforward outputs, and bands show 95\% image-bootstrap confidence intervals.}
    \label{fig:continue_optimization}
\end{figure}

\paragraph{Continued optimization.}
Figure~\ref{fig:continue_optimization} evaluates the predicted representations as initializations for further per-image fitting.
Gaussifier uses our fitting pipeline (Appendix~\ref{app:continued_optimization}), while Instant-GI and Fast-2DGS use their own native refinement procedures.
For AIR, we implement a refinement extension with its Predict-Optimize-Distill parameterization and optimization rules.
At 0.2\,s, Gaussifier reaches 42.94/36.26\,dB on Kodak-24/DIV2K. At 20\,s, its quality rises to 46.79/40.62\,dB.
Gaussifier is therefore also a good starting point for further fitting. It also outperforms all three baselines at every measured time up to 20\,s.

\begin{figure}[!htbp]
  \centering
  \includegraphics[width=\linewidth]{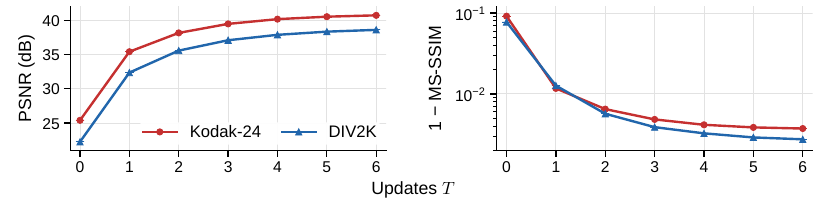}
  \caption{\textbf{Quality after each update.} Feedforward reconstruction quality improves as $T$ increases and gradually plateaus beyond $T=4$. $1-\mathrm{MS\text{-}SSIM}$ is shown on a logarithmic scale.}
  \label{fig:ablation_correction_steps}
\end{figure}

\begin{table}[!htbp]
\centering
\caption{Same-count comparison with the released baselines. Gaussifier uses the per-image count of each baseline, no fitting, and at most 14\,ms. Base and Ours are PSNR in dB. Fit is the time a baseline needs with its own fitting to reach our PSNR, with the ratio to 14\,ms. Fast-2DGS uses 50k Gaussians. Tables~\ref{tab:matched} and~\ref{tab:matched_fitting} add AIR ps6/ps7.}
\label{tab:same_count}
\setlength{\tabcolsep}{3.5pt}
\begin{tabular}{@{}lrrrrcrrrr@{}}
\toprule
& \multicolumn{4}{c}{Kodak-24} && \multicolumn{4}{c}{DIV2K} \\
\cmidrule(lr){2-5}\cmidrule(lr){7-10}
Baseline & \#G (k) & Base & Ours & Fit && \#G (k) & Base & Ours & Fit \\
\midrule
AIR        & 35.4 & 31.86 & \textbf{39.86} & 0.81\,s (58$\times$)  && 36.3 & 27.57 & \textbf{34.90} & 1.21\,s (86$\times$) \\
Instant-GI & 25.1 & 30.40 & \textbf{37.83} & 0.87\,s (62$\times$)  && 28.8 & 27.41 & \textbf{33.11} & 0.60\,s (43$\times$) \\
Fast-2DGS  & 50.0 & 25.95 & \textbf{41.56} & 4.26\,s (304$\times$) && 50.0 & 23.29 & \textbf{37.07} & 4.56\,s (326$\times$) \\
\bottomrule
\end{tabular}
\end{table}

\subsection{Ablation Studies}
\label{sec:ablation_studies}

\paragraph{Refinement depth.}

Figure~\ref{fig:ablation_correction_steps} shows that average reconstruction quality improves after every update.
PSNR rises from 25.40 to 40.73\,dB on Kodak-24 and from 22.25 to 38.61\,dB on DIV2K.
These are total gains of 15.33/16.36\,dB over $T=0$.
The gains shrink at later steps, but the last update still adds 0.20/0.26\,dB.
All of these improvements come from repeatedly updating the same set of Gaussians with a fixed count.
Beyond the six trained updates, PSNR still rises by 0.15/0.31\,dB, peaking at eight and ten updates, and then slowly declines.
Adam started from the same initial $T=0$ Gaussians needs a median of 110/77 steps to reach the PSNR of six updates (Appendix~\ref{app:refinement_checks}).

\paragraph{Density maps.}

\begin{table}[t]
\centering
\caption{Component ablations. The density and color rows replace one predicted map at test time. The $\boldsymbol U_{\boldsymbol\phi}$ rows retrain $\boldsymbol U_{\boldsymbol\phi}$ without the listed inputs.}
\label{tab:ablation_components}

\setlength{\tabcolsep}{5pt}
\renewcommand{\arraystretch}{1.06}

\begin{tabular}{@{}lrrrr@{}}
\toprule
& \multicolumn{2}{c}{Kodak-24}
& \multicolumn{2}{c}{DIV2K} \\
\cmidrule(lr){2-3}
\cmidrule(lr){4-5}
Variant
& PSNR $\uparrow$
& MS-SSIM $\uparrow$
& PSNR $\uparrow$
& MS-SSIM $\uparrow$ \\
\midrule

\textbf{Gaussifier (Full)}
& \textbf{40.73}
& \textbf{0.9963}
& \textbf{38.61}
& \textbf{0.9973} \\

\addlinespace[4pt]

w/o predicted density (uniform)
& 34.86
& 0.9875
& 32.48
& 0.9853 \\

w/o predicted density (gradient)
& 39.45
& 0.9956
& 36.81
& 0.9961 \\

\addlinespace[4pt]

w/o predicted RGB initialization
& 31.81
& 0.9830
& 28.82
& 0.9803 \\

\addlinespace[4pt]
w/o density and count inputs in $\boldsymbol U_{\boldsymbol\phi}$
& 40.39
& 0.9953
& 38.31
& 0.9969 \\

w/o initial attribute inputs in $\boldsymbol U_{\boldsymbol\phi}$
& 40.54
& 0.9960
& 38.51
& 0.9973 \\

\bottomrule
\end{tabular}

\end{table}

We compare the predicted density with uniform and gradient-guided alternatives.
The gradient-guided density follows Image-GS~\citep{zhang2025imagegs}, which mixes normalized grayscale Sobel magnitudes with a uniform distribution, using weights 0.7 and 0.3.
We keep EMV, the per-image Gaussian count, and the predicted attribute maps unchanged.
Table~\ref{tab:ablation_components} shows that replacing the density with a uniform one loses 5.87/6.13\,dB on Kodak-24/DIV2K, and with the gradient-based density loses 1.28/1.80\,dB. The trained pipeline therefore relies on the learned density.

\paragraph{Color initialization.}
We also replace the predicted colors with the input-image RGB, sampled bilinearly at the Gaussian centers, and keep the predicted covariances and density unchanged.
This yields 31.81/28.82\,dB on Kodak-24/DIV2K after all updates.
The trained refinement therefore relies on the predicted color coefficients (Section~\ref{subsec:training_objectives_and_schedule}).

\paragraph{Refinement network inputs.}
We examine the input context of $\boldsymbol U_{\boldsymbol\phi}$ by removing either the density and count maps or the initial log-covariance and RGB maps, and retraining the refinement network.
Removing the density and count maps reduces PSNR by 0.34/0.30\,dB on Kodak-24/DIV2K (Table~\ref{tab:ablation_components}).
These maps help the network relate predicted representation demand to the current spatial distribution of Gaussian centers.
Removing the initial attribute maps reduces PSNR by 0.19/0.10\,dB.
Since the effect of a parameter update depends on a Gaussian's covariance and color coefficients, the initial attribute maps help the network choose appropriate update magnitudes for a given reconstruction residual.
Both input groups provide useful context for refinement, with modest gains in PSNR.

\paragraph{Density samplers.}

\begin{table}[!htbp]
\centering
\caption{Density sampler ablation at two image resolutions. We report RGB PSNR (dB) after $T=6$ updates, while keeping the trained weights, density, and target number of Gaussians fixed. The weights were fine-tuned with EMV.}
\label{tab:ablation_samplers}
\setlength{\tabcolsep}{6pt}
\renewcommand{\arraystretch}{1.06}
\begin{tabular}{@{}lrrrr@{}}
\toprule
& \multicolumn{2}{c}{Kodak-24}
& \multicolumn{2}{c}{DIV2K} \\
\cmidrule(lr){2-3}\cmidrule(lr){4-5}
Sampler
& $512\!\times\!512$
& $1024\!\times\!1024$
& $512\!\times\!512$
& $1024\!\times\!1024$ \\
\midrule
Equal-Mass Voronoi (\textbf{ours})  & \textbf{40.73} & \textbf{46.79} & 38.61 & \textbf{41.25} \\
Hilbert-Lloyd       & 40.25 & 46.71 & \textbf{38.64} & 40.59 \\
Hilbert Equal-Mass  & 39.81 & 42.85 & 37.15 & 35.59 \\
Hilbert Inverse-CDF & 40.37 & 45.53 & 38.40 & 40.67 \\
Error-Diffusion     & 38.14 & 40.23 & 36.41 & 38.02 \\
Poisson             & 35.26 & 35.67 & 33.66 & 34.69 \\
Binomial            & 35.36 & 35.74 & 33.83 & 34.86 \\
\bottomrule
\end{tabular}
\end{table}

Table~\ref{tab:ablation_samplers} compares seven sampling strategies at $512\times512$ and $1024\times1024$.
EMV achieves the highest PSNR in three of the four settings, trailing Hilbert-Lloyd by only 0.03\,dB on DIV2K at $512\times512$.
Relative to Error-Diffusion, EMV improves PSNR by 2.59/2.20\,dB on Kodak-24/DIV2K at $512\times512$, with larger gains of 6.56/3.23\,dB at $1024\times1024$.
Poisson and binomial sampling perform substantially worse across all settings.
These results support EMV as an effective placement within our trained pipeline.

\section{Limitations and discussion}
\label{sec:limitations_and_discussion}
We evaluate Gaussifier mainly at one image resolution and report speed on one GPU.
The cost grows with the image resolution because the refinement network and the renderer run at full resolution in every update.
Gaussifier runs at 20 FPS at $1280\times720$ (Appendix~\ref{app:evaluation_details}).
Compiled, AIR and Fast-2DGS run as fast as or faster than Gaussifier (Appendix~\ref{app:evaluation_details}).
However, at the same Gaussian count their outputs are 7 to 16\,dB lower (Table~\ref{tab:same_count}).
Without the predicted density, the same refinement loses about 6\,dB (Table~\ref{tab:ablation_components}).
The network is trained with six updates, and quality slowly declines after eight to ten steps (Appendix~\ref{app:refinement_checks}).
Gaussifier does not replace long per-image fitting.
On Kodak-24, 20\,s of fitting from its output adds another 6\,dB (Figure~\ref{fig:continue_optimization}).
Training with more updates could close part of this gap.
Future work could add video-specific improvements and fine-tuning that exploit similar adjacent frames.
For example, Gaussians could carry over between frames.
Other directions are training at higher resolutions and feeding Gaussifier outputs to downstream vision models.

\section{Conclusion}
\label{sec:conclusion}
Gaussifier shows that high-quality 2D Gaussian decomposition does not require per-image optimization.
At the same Gaussian count, it reaches in milliseconds the quality that per-image fitting reaches in seconds.
Where to place the Gaussians and how to correct them can both be learned from many images.
The Gaussian count and the number of updates then trade speed for quality within one trained model rather than through an optimization run.
2D Gaussians can therefore serve as a representation that real-time systems produce on demand.
The same recipe of learning placement from per-image fits and learning corrections across images could also apply to 3D Gaussians.

\bibliography{references}
\bibliographystyle{gaussifier}
\clearpage
\appendix
\section{Method Details}
\label{app:method_details}

\subsection{Architecture and Parameterization}

The image backbone is a residual U-Net with base width 32 and four resolution levels.
Three $1\times1$ prediction heads follow it.
The heads output one log-density channel, three independent entries of a symmetric log-covariance matrix, and three sigmoid-activated RGB channels.
The refinement U-Net has two parallel branches with base widths 32 and 24.
Each branch has two downsampling stages and receives the same context.
Their outputs are summed.
Each branch predicts eight update channels, with three for log-covariance, three for RGB, and two for position.
The complete model has 2,802,895 parameters, and 746,792 of them belong to the refinement module.

The refinement input has 17 channels.
They are the input RGB image, the mean-normalized density, the initial log-covariance map, the initial RGB map, the current RGB reconstruction, the RGB residual, and a single-channel count raster of the centers.
Each center adds one count to the pixel that contains it.
This raster is recomputed after position updates.
The initial attribute maps and the cell-mass weights remain fixed.
The evolving Gaussian parameters and their rendering carry the information between steps.

The main text uses pixel coordinates for spatial quantities.
The implementation stores centers in $[0,1]^2$ and covariances in pixel units, and it converts the centers before rasterization.
For a raw two-channel position prediction $\boldsymbol d_i^t$, the normalized displacement is $2.56\boldsymbol d_i^t$ divided by the longer image side.
The increment $\Delta\boldsymbol\mu_i^t$ in Eq.~\ref{eq:parameter_updates} includes this fixed coordinate scaling.
Covariance updates act on the three independent entries of the symmetric matrix before exponentiation.
The density equations in the main text omit numerical clipping and small stability constants.
Count prediction evaluates the exponential after clipping the logits to $[-20,8]$.
Training caps the count at $N_{\max}=\min(10^6,\lfloor0.4P\rfloor)$.
Evaluation uses the predicted count without this cap.

\subsection{Approximate Equal-Mass Placement}
\label{app:emv_details}

The initial sites come from stratified sampling over randomly shifted tiles.
Each tile receives a share of the sites proportional to its density mass.
Error diffusion over the tiles rounds these shares to integers.
Inside a tile, sites are drawn at distinct pixels in proportion to $\rho$ and jittered within the pixel.
Placement is deterministic for a fixed seed.
Across seeds 0, 1, and 2, the mean PSNR after six updates has a standard deviation of 0.007 and 0.001\,dB on Kodak-24 and DIV2K.
Each merge round removes at most a third of the sites.
Competing proposals to the same target are resolved by source mass, and a losing source tries its next available neighbor.
Jump flooding~\citep{rong2006jfa} gives approximate nearest-center assignments on the pixel grid.
Candidates with no assigned mass are removed before matching.
Additional merges enforce the requested final count.
Lloyd updates alone do not impose equal capacities.
The training code checks for degenerate cell configurations.
In that case, it can fall back to an equal-mass Hilbert partition followed by Lloyd updates.
The final training pass promoted checkpoints only when no fallback occurred, and the released inference runtime has no fallback.
On Kodak-24 and DIV2K, 0.002\% of the cells are empty and 0.001\% of the centers are duplicates.
On an RTX 5090, placement takes 1.84 and 2.53\,ms on Kodak-24 and DIV2K, and 2.06 and 2.75\,ms with the cell-mass weights.

EMV reduces the spread of the cell masses but does not make them exactly equal.
The coefficient of variation of $m_i/\bar m$ is 0.47 and 0.46 on Kodak-24 and DIV2K when $N$ sites are drawn directly by stratified sampling.
It is 0.31 and 0.35 after merging, and 0.27 and 0.32 after the three Lloyd iterations.
Cells average 6.2 and 3.6 pixels, so pixel quantization adds part of this spread.

The final cell assignments are used to compute the cell-mass weights $w_i$.
These assignments are fixed in the backward pass.
However, the density accumulation within each cell remains differentiable.
The weights therefore provide a gradient for the shape of the normalized density.
They do not differentiate the integer count or the movement of cell boundaries.
They are also invariant to a global rescaling of the density when the count and assignments are fixed.
Absolute count supervision and the calibration in Appendix~\ref{app:count_calibration} provide the complementary budget signal.
Setting all weights to one at inference lowers the initial PSNR by 3.9 and 2.8\,dB on Kodak-24 and DIV2K.
After six updates, the loss shrinks to 0.22 and 0.29\,dB.

\subsection{Fixed Gaussian Budgets}
\label{app:fixed_budgets}
The density head predicts an image-specific count $N_{\mathrm{auto}}$, and the covariance map is learned for that count.
A user can instead request a budget $N$.
EMV then places $N$ centers from the same normalized density $\rho$.
Rendering is additive (Eq.~\ref{eq:gaussian_render}).
Pinning $N$ above $N_{\mathrm{auto}}$ without adjusting the covariances therefore brightens the initial rendering roughly in proportion to $N/N_{\mathrm{auto}}$.
Far above the predicted count, most pixels saturate.
For $N>N_{\mathrm{auto}}$, we therefore scale every initial covariance by $N_{\mathrm{auto}}/N$.
We add $\log(N_{\mathrm{auto}}/N)$ to the diagonal of $\boldsymbol S_i^0$, which keeps the total area covered by the Gaussians constant.
Below $N_{\mathrm{auto}}$, the covariances are left unchanged.
Refinement recovers the dimmer initial rendering.
Enlarging the covariances there changed the final PSNR by $-0.33$ to $+0.40$\,dB, depending on the count.
The rule applies only at inference, adds no parameters, and does not affect the adaptive-count results.

\section{Training Details}
\label{app:training_stages}

\subsection{Reference Fits}
\label{app:reference_fits}
We use two sets of Open Images data.
Open Images is not an evaluation dataset in this work.
The first set has about 107k images from the Open Images test split.
They are fitted for 2,800 steps with our fitting pipeline.
It optimizes the centers, scales, rotations, and colors of all Gaussians with Adam on the MSE through the additive renderer.
Its adaptive density control periodically moves Gaussians that leave the image, cover no pixels, or grow very large to pixels with high error.
A ridge regression on patch gradient, Laplacian, and color-variance statistics sets the initial count.
Images below 38\,dB are refit once with twice the Gaussians, and the better fit is kept.
These fits have a median of 64k Gaussians and 42.3\,dB PSNR.
Their optimization took 326 GPU-hours in total.
The second set has about 20k images from the Open Images train split.
Their fits start from an earlier Gaussifier prediction and are optimized for 2,400 steps.
These fits took about 33 GPU-hours.

\subsection{Training Stages}
\label{app:stages}
The released model comes from a sequence of runs.
Pretraining on the first set uses the density loss and the covariance and color losses with weights 0.1 and 1.0.
It adds a small reconstruction loss in its last epochs.
The refinement network is then trained on the same images with one, two, and then four updates.
These runs first use the MSE and then an $L_1$ and SSIM loss.
Some of them also update the U-Net.
Some add losses toward the render and the centers refined by 100 Adam steps from the current Gaussians.
The U-Net is then fine-tuned on the second set with the losses of Section~\ref{subsec:training_objectives_and_schedule} and weights 0.05 and 0.1.
The refinement network continues on the second set and then on images alone, with four and then six updates.
Three later stages update the output heads and the refinement network.
One of them also updates the backbone at a learning rate of $3\times10^{-8}$.
These stages move each center to the centroid of its cell before rendering and pass gradients to the density through soft centroids.
Their count losses raise the predicted count on images with low PSNR and lower it on images with very high PSNR.
Another loss moves density toward tiles with large residuals.
The next stage adapts the output heads to Hilbert-Lloyd placement (Appendix~\ref{app:sampler_ablation}).
The refinement network is then trained with the increasing and the final-state MSE, and the two results are averaged with weights 0.75 and 0.25.
Count calibration (Appendix~\ref{app:count_calibration}) and the final adaptation (Appendix~\ref{app:final_adaptation}) follow.
These stages add no networks or count search at inference.
Table~\ref{tab:training_compute} lists the stages and their wall-clock time.
\begin{table}[!htbp]
\centering
\caption{Training stages of the released model and their wall-clock time on one RTX 5090. The times include the steps after the retained checkpoints.}
\label{tab:training_compute}
\small
\setlength{\tabcolsep}{4pt}
\begin{tabular}{@{}lllr@{}}
\toprule
Stage & Data & Trained parts & Hours \\
\midrule
Pretraining & First set & U-Net & 4.3 \\
Refinement with one to four updates & First set & Refinement, partly the U-Net & 47.9 \\
Fine-tuning and paired refinement & Second set & U-Net, then refinement & 27.7 \\
Refinement with four, then six updates & Images only & Refinement & 153.4 \\
Centroid, count, and local-density losses & Images only & Heads, refinement, briefly the backbone & 20.8 \\
Hilbert-Lloyd adaptation and MSE catch-up & Images only & Heads, then refinement & 6.5 \\
Count calibration & 4,096 images & Density head & $<$1 \\
Final adaptation & Images only & Heads and refinement & 166.6 \\
\midrule
Total & & & 428 \\
\bottomrule
\end{tabular}
\end{table}

\subsection{Count Calibration}
\label{app:count_calibration}
We calibrate the density head on 4,096 training images with a small discrete search around the predicted budget of each image.
The candidate multipliers are $0.9$, $1.0$, and $1.1$.
Among the candidates that reach a 41\,dB quality threshold, we select the smallest count.
If none reaches the threshold, we select the candidate with the highest PSNR.
An increase is accepted only when it improves the measured reconstruction quality over the original budget.
These targets are local and empirically evaluated.
We do not claim that they are globally optimal Gaussian counts.

Let $N^\star$ denote a selected target and $N_\theta=\sum_{\boldsymbol x\in\Omega}\lambda_\theta(\boldsymbol x)$ the predicted count.
Let $p_\theta=\rho_\theta/P$ and $p_{\mathrm{ref}}$ be the predicted and pre-calibration spatial distributions.
The calibration objective is
\begin{equation}
    \cL_{\mathrm{cal}}=
    (\log N_\theta-\log N^\star)^2
    +\beta D_{\mathrm{KL}}(p_{\mathrm{ref}}\,\|\,p_\theta),
    \qquad \beta=4.
    \label{eq:count_calibration}
\end{equation}
For examples with an unchanged budget, the regression target keeps the original unrounded density sum.
This prevents count caps from causing a systematic downward bias.
Only the 33 parameters of the density head are updated in this stage.
At inference, the sum of the predicted density gives the count directly.

\subsection{Final Adaptation}
\label{app:final_adaptation}
The final stage uses a batch size of two and the increasing-state MSE objective in Eq.~\ref{eq:reconstruction_loss}.
The image backbone is frozen.
AdamW updates the refinement module at an initial learning rate of $5\times10^{-8}$.
It updates the density and attribute heads at one tenth of this rate, with zero weight decay.
A cosine schedule uses 5,000 warmup steps and a minimum refinement learning rate of $5\times10^{-9}$.
We select the retained checkpoint at step 540,000, after 1,080,000 image presentations.
The selection uses 256 images from the validation part of the 99:1 split in Section~\ref{subsec:training_objectives_and_schedule}.
Its criterion emphasizes the lower tail of reconstruction quality.
No Kodak-24 or DIV2K image is used for checkpoint selection or count calibration.
Its sampler uses hard positions and assignments, and density gradients pass through the cell-mass weights.

\section{Evaluation Details}
\label{app:evaluation_details}

\subsection{Protocol}
All timings use one NVIDIA RTX 5090 with batch size one.
Gaussifier runs its C++ deployment with a TensorRT FP16 U-Net, an FP16 refinement network compiled with PyTorch AOTInductor (AOTI), fused update kernels, and seed 12345.
Each image runs three warm-up passes and 30 timed passes with synchronized wall-clock timing.
We take the median time of each image and average it over the dataset.
FPS is 1,000 divided by this average in milliseconds.
Memory in Table~\ref{tab:main_table} is the peak device memory of each process, as reported by NVML.
It includes the CUDA context and the memory of every library.
AIR and Instant-GI use seed 0.
For Fast-2DGS, quality is averaged over seeds 0, 1, and 2, and timing uses seed 0.
The baselines run their released code in eager PyTorch FP32.
Table~\ref{tab:efficiency_details} adds latency percentiles, eager Gaussifier, and the peak of the PyTorch or LibTorch allocator.
The allocator peak excludes the CUDA context and memory managed by TensorRT or CuPy.
In eager PyTorch FP32, Gaussifier takes 49.6/50.6\,ms and is about as fast as AIR.
Table~\ref{tab:resolution} reports Gaussifier on DIV2K at $1280\times720$ and $1920\times1080$.
Its cost grows with the number of pixels and Gaussians.
It runs at 20 and 8\,FPS at these sizes.
\begin{table}[!htbp]
\centering
\caption{Runtime details at $512\times512$ on an RTX 5090. Latency is the median of 30 timed passes for each image, summarized over the images by the mean, the 95th percentile, and the maximum. Allocated memory is the peak of the PyTorch or LibTorch allocator. Device memory is the NVML peak of the process, as in Table~\ref{tab:main_table}. Memory is given as Kodak-24/DIV2K. The compiled baselines keep their released checkpoints and use torch.compile with CUDA graphs in FP16, or in BF16 for Instant-GI, whose activations overflow FP16. The rasterizers and the CuPy triangulation of Instant-GI stay as released. Compilation changes their PSNR by at most 0.01\,dB. CUDA graphs hide part of their allocation, so we omit it.}
\label{tab:efficiency_details}
\footnotesize
\setlength{\tabcolsep}{3.5pt}
\begin{tabular}{@{}llccrr@{}}
\toprule
& & \multicolumn{2}{c}{Latency (ms), mean / p95 / max} & & \\
\cmidrule(lr){3-4}
Method & Runtime & Kodak-24 & DIV2K & Allocated MB & Device MB \\
\midrule
Gaussifier & TensorRT/AOTI FP16 & 13.7 / 14.5 / 14.9 & 14.7 / 16.4 / 23.8 & 180/183 & 1,126/1,130 \\
Gaussifier & PyTorch FP32 & 49.6 / 50.8 / 50.9 & 50.6 / 52.2 / 57.4 & 448/449 & 1,304/1,306 \\
AIR & PyTorch FP32 & 53.0 / 53.2 / 53.2 & 53.2 / 53.3 / 53.4 & 1,912/1,912 & 2,852/2,852 \\
Instant-GI & PyTorch FP32, CuPy & 65.4 / 75.4 / 79.2 & 68.7 / 80.6 / 84.7 & 999/999 & 3,364/3,370 \\
Fast-2DGS & PyTorch FP32 & 6.3 / 6.3 / 6.4 & 6.4 / 6.4 / 6.5 & 310/310 & 1,022/1,022 \\
\midrule
AIR & compiled FP16 & 12.3 / 12.4 / 12.4 & 12.4 / 12.4 / 12.4 & -- & 2,460/2,460 \\
Instant-GI & compiled BF16, CuPy & 31.6 / 36.9 / 38.4 & 32.4 / 38.6 / 41.5 & -- & 3,010/3,364 \\
Fast-2DGS & compiled FP16 & 1.7 / 1.7 / 1.7 & 1.8 / 1.8 / 1.9 & -- & 1,224/1,238 \\
\bottomrule
\end{tabular}
\end{table}

\begin{table}[!htbp]
\centering
\caption{Gaussifier on DIV2K at larger frame sizes on an RTX 5090, with the TensorRT and AOTI modules built for each size. Each image is resized to cover the frame and center-cropped. Latency and memory are measured as in Table~\ref{tab:efficiency_details}.}
\label{tab:resolution}
\footnotesize
\setlength{\tabcolsep}{4pt}
\begin{tabular}{@{}lrcrcrr@{}}
\toprule
Frame size & \#G (k) & Latency (ms), mean / p95 / max & FPS & PSNR & Allocated MB & Device MB \\
\midrule
$512\times512$ & 72.2 & 14.7 / 16.4 / 23.8 & 68 & 38.61 & 183 & 1,130 \\
$1280\times720$ & 163.0 & 51.1 / 59.1 / 75.8 & 20 & 40.35 & 612 & 2,264 \\
$1920\times1080$ & 243.1 & 125.2 / 133.9 / 149.8 & 8 & 42.01 & 1,348 & 4,132 \\
\bottomrule
\end{tabular}
\end{table}

\subsection{Baselines}
\label{app:baselines}
AIR uses its released final full-precision stage with codec execution disabled.
AIR releases checkpoints for patch sizes 5, 6, and 7 (ps5, ps6, and ps7).
Smaller patches give more Gaussians.
Table~\ref{tab:main_table} reports ps5, and Table~\ref{tab:matched} also reports ps6 and ps7.
Instant-GI uses its released checkpoint with kernel size 3 (ks3).
This setting max-pools the position map over $3\times3$-pixel blocks before dithering.
Instant-GI runs with its CuPy Delaunay and native dithering and ellipse extensions.
Fast-2DGS uses its released heatmap-prior checkpoint and its attribute checkpoint with offset refinement (Plus).
It places 50k Gaussians on Kodak-24 and 70k on DIV2K.
Instant-GI$^*$ and Fast-2DGS$^*$ continue training from these official checkpoints on the Open Images training images of Section~\ref{subsec:training_objectives_and_schedule}.

\subsection{Sampler Ablation}
\label{app:sampler_ablation}
Table~\ref{tab:ablation_samplers} keeps the trained weights, the density, and the count fixed and changes only the placement.
The cell-mass weights are computed in the same way for every sampler.
Hilbert Equal-Mass orders the pixels along a Hilbert curve~\citep{hilbert1891} and cuts the cumulative mass into $N$ equal segments.
It places a center at the mass centroid of each segment.
Hilbert-Lloyd adds three Lloyd iterations to these centers.
Hilbert Inverse-CDF places the $N$ centers by inverting the cumulative distribution function (CDF) of the density along the same curve.
Error-Diffusion draws $N$ centers directly with the stratified sampler of Appendix~\ref{app:emv_details}, without merging or Lloyd iterations.
Poisson and Binomial draw a count for each pixel, with eight trials for Binomial, and jitter the points within the pixel.
Their total count therefore varies around $N$.

\section{Additional Results}
\label{app:additional_results}

\subsection{Refinement Checks}
\label{app:refinement_checks}
These checks follow the protocol of Table~\ref{tab:main_table} in eager FP32 PyTorch with seed 12345 and the production cell-mass weights.
With the same weights, PSNR after 6, 8, 10, and 12 updates is 40.74, 40.89, 40.79, and 40.55\,dB on Kodak-24.
On DIV2K, it is 38.61, 38.87, 38.92, and 38.82\,dB.

For the Adam comparison, Adam starts from the initial Gaussians of the same run.
It optimizes the same centers, log-covariances, and colors with the same renderer, weights, and projections as Eq.~\ref{eq:parameter_updates}.
It uses the MSE and constant learning rates.
The learning rates are $4\times10^{-4}$ for the normalized centers and 0.02 for the log-covariances and colors, the best of six scales on four Kodak-24 and four DIV2K images.
Tuning on test images favors Adam.
Six learned updates take 42\,ms in this implementation, and one Adam step takes about 0.9 to 1.0\,ms.
At equal time, Adam is 2.1 and 2.5\,dB lower on Kodak-24 and DIV2K.
Adam reaches the PSNR of six updates after a median of 110 steps on Kodak-24 and 77 steps on DIV2K.
Four DIV2K images do not reach it within 1,000 steps.
After 1,000 steps, Adam reaches 44.40 and 43.87\,dB.

\subsection{GaussianImage at a Matched Count}
\label{app:gaussianimage_matched}
This comparison supports the Gaussian budget result in Section~\ref{subsec:feedforward_reconstruction}.
We fit GaussianImage~\citep{zhang2024gaussianimage} to all 24 Kodak images with the preprocessing of Section~\ref{subsec:experimental_setup}.
It starts from its native initialization, and its schedule is adjusted to a final budget of 50k Gaussians.
After 20\,s of fitting on an NVIDIA L40S GPU, it reaches 41.37\,dB.
Gaussifier reaches 41.56\,dB with the same 50k Gaussians and no fitting.

\subsection{Continued Optimization}
\label{app:continued_optimization}
For Figure~\ref{fig:continue_optimization}, our fitting pipeline (Appendix~\ref{app:reference_fits}) starts from the final Gaussifier output.
On DIV2K, it starts from Gaussifier at the lowest per-image count among the three baselines, 28.8k on average.
This count is below the predicted count on 98 of the 100 images, where the covariances stay unchanged (Appendix~\ref{app:fixed_budgets}).
Its adaptive density control is disabled.
Its learning rates warm up for 25 steps and then decay exponentially over a 50k-step horizon.
The reference fits in Figure~\ref{fig:feedforward_reconstruction_quality_and_Gaussian_geometry} use the same setting for 50k steps.
For AIR, we apply its Predict-Optimize-Distill (POD) optimizer to all accumulated Gaussians at once.
This extension keeps the Adam optimizer, the learning rate of 0.001, and the $0.7L_1+0.3(1-\mathrm{SSIM})$ loss of AIR.
Instant-GI and Fast-2DGS use their released fitting code with the Adan and AdamW optimizers.
Table~\ref{tab:budget_psnr} lists the values behind Figure~\ref{fig:continue_optimization}.
\begin{table}[!htbp]
\centering
\caption{PSNR (dB) after a given total elapsed time per image on an RTX 5090. The time includes initialization, the transfer to the optimizer, and per-image fitting, and the runs are those of Figure~\ref{fig:continue_optimization}, plus AIR ps6 and ps7 for Table~\ref{tab:matched_fitting}. Gaussifier is refined with our fitting pipeline, with adaptive density control disabled. Instant-GI and Fast-2DGS use their native refinement, and AIR uses our POD-based refinement extension. A dash marks budgets at which a pipeline has not yet produced a result for every image. On DIV2K, Gaussifier fits at the lowest per-image count among AIR, Instant-GI, and Fast-2DGS (Appendix~\ref{app:continued_optimization}). Without any fitting, Gaussifier reaches 40.73\,dB on Kodak in 13.7\,ms (Table~\ref{tab:main_table}), and the best competing pipeline stays below this until after 1\,s.}
\label{tab:budget_psnr}
\setlength{\tabcolsep}{5pt}
\renewcommand{\arraystretch}{1.15}
\begin{tabular}{lcccccccccc}
\toprule
Pipeline & 0.02\,s & 0.05\,s & 0.1\,s & 0.2\,s & 0.5\,s & 1\,s & 2\,s & 5\,s & 10\,s & 20\,s \\
\midrule
\multicolumn{11}{l}{\textit{Kodak-24}} \\
Gaussifier & \textbf{40.82} & \textbf{41.13} & \textbf{42.28} & \textbf{42.94} & \textbf{43.79} & \textbf{44.43} & \textbf{45.09} & \textbf{45.85} & \textbf{46.33} & \textbf{46.79} \\
AIR ps5    & --    & --    & 34.32 & 36.26 & 38.69 & 40.38 & 41.85 & 43.32 & 44.14 & 44.75 \\
AIR ps6    & --    & --    & 33.30 & 34.80 & 36.64 & 37.96 & 39.11 & 40.24 & 40.80 & 41.17 \\
AIR ps7    & --    & --    & 32.55 & 33.73 & 35.17 & 36.22 & 37.14 & 38.00 & 38.41 & 38.66 \\
Instant-GI & --    & --    & 33.04 & 34.50 & 36.65 & 38.12 & 39.22 & 40.35 & 41.07 & 41.53 \\
Fast-2DGS  & 28.70 & 30.80 & 32.13 & 33.55 & 35.79 & 37.53 & 39.95 & 41.91 & 42.72 & 43.04 \\
\midrule
\multicolumn{11}{l}{\textit{DIV2K}} \\
Gaussifier & \textbf{33.40} & \textbf{34.80} & \textbf{35.50} & \textbf{36.26} & \textbf{37.37} & \textbf{38.21} & \textbf{38.94} & \textbf{39.72} & \textbf{40.21} & \textbf{40.62} \\
AIR ps5    & --    & --    & 29.25 & 30.80 & 32.86 & 34.47 & 36.01 & 37.62 & 38.48 & 39.10 \\
AIR ps6    & --    & --    & 28.57 & 29.75 & 31.25 & 32.40 & 33.47 & 34.53 & 35.03 & 35.34 \\
AIR ps7    & --    & --    & 28.05 & 28.99 & 30.16 & 31.05 & 31.87 & 32.62 & 32.93 & 33.10 \\
Instant-GI & --    & --    & 29.46 & 30.75 & 32.69 & 34.26 & 35.52 & 36.69 & 37.35 & 37.79 \\
Fast-2DGS  & 25.74 & 27.51 & 28.55 & 29.68 & 31.67 & 33.33 & 35.61 & 37.24 & 37.83 & 38.08 \\
\bottomrule
\end{tabular}
\end{table}

\subsection{Matched Comparisons}
Tables~\ref{tab:matched} and~\ref{tab:matched_fitting} give the per-baseline results of the controlled comparisons in Section~\ref{subsec:feedforward_reconstruction}.
\begin{table}[!htbp]
\centering
\caption{Matched feedforward comparisons with the released baselines at the same Gaussian count. We re-ran all methods on the same RTX 5090 under the protocol of Table~\ref{tab:main_table}. Fast-2DGS uses 50k Gaussians on Kodak and 70k on DIV2K. Gaussifier uses the per-image count of each baseline. In the \textbf{Same count} column, it runs all $T{=}6$ updates, and it has the higher PSNR on every image. In the \textbf{Same count and PSNR} columns, it uses the fewest updates whose mean PSNR reaches that of the baseline. This is $T{=}1$, and $T{=}2$ for AIR ps7 on DIV2K. Since $T$ is discrete, the PSNR reached can exceed that of the baseline. For Fast-2DGS, one update already exceeds its PSNR by 8.3 to 9.5\,dB, so we do not compare times. \#G is the number of Gaussians. Times are batch-one synchronized wall times in ms, given as the mean of per-image medians over 30 repetitions. Gaussifier runs in its TensorRT/AOTI deployment, and the baselines run in their compiled implementations (Table~\ref{tab:efficiency_details}). Gaussifier's PSNR comes from its eager PyTorch implementation, which matches the deployment within 0.02\,dB. At the fixed counts of Fast-2DGS, it comes from the deployment with the covariance rule of Appendix~\ref{app:fixed_budgets}.}
\label{tab:matched}
\setlength{\tabcolsep}{4pt}
\renewcommand{\arraystretch}{1.15}
\begin{tabular}{@{}lrrrccc@{}}
\toprule
& \multicolumn{3}{c}{Baseline} & Same count & \multicolumn{2}{c}{Same count and PSNR} \\
\cmidrule(lr){2-4}\cmidrule(lr){5-5}\cmidrule(lr){6-7}
Baseline & \#G & ms & PSNR & Our PSNR & Our ms & Our PSNR \\
\midrule
\multicolumn{7}{@{}l}{\textit{Kodak-24}} \\
AIR ps5    & 35,386 & 12.3 & 31.86 & 39.86 (+8.0)  & 4.65 (2.7$\times$) & 34.75 \\
AIR ps6    & 25,468 & 12.4 & 31.16 & 37.84 (+6.7)  & 4.41 (2.8$\times$) & 32.94 \\
AIR ps7    & 19,260 & 12.4 & 30.60 & 35.70 (+5.1)  & 4.27 (2.9$\times$) & 30.65 \\
Instant-GI & 25,126 & 31.6 & 30.40 & 37.83 (+7.4)  & 4.45 (7.1$\times$) & 32.83 \\
Fast-2DGS  & 50,000 & 1.7  & 25.95 & 41.56 (+15.6) & --                 & --    \\
\midrule
\multicolumn{7}{@{}l}{\textit{DIV2K}} \\
AIR ps5    & 36,294 & 12.4 & 27.57 & 34.90 (+7.3)  & 4.62 (2.7$\times$) & 29.59 \\
AIR ps6    & 25,706 & 12.4 & 27.10 & 32.14 (+5.0)  & 4.38 (2.8$\times$) & 27.25 \\
AIR ps7    & 19,260 & 12.4 & 26.71 & 29.74 (+3.0)  & 6.08 (2.0$\times$) & 28.16 \\
Instant-GI & 28,825 & 32.4 & 27.41 & 33.11 (+5.7)  & 4.46 (7.3$\times$) & 28.06 \\
Fast-2DGS  & 70,000 & 1.8  & 23.73 & 38.78 (+15.1) & --                 & --    \\
\bottomrule
\end{tabular}
\end{table}

\begin{table}[!htbp]
\centering
\caption{Baselines with their own per-image fitting versus Gaussifier without fitting, at the same Gaussian count. Each baseline starts from its released feedforward prediction and applies its native refinement, which keeps its count fixed. For AIR, we use our POD-based extension. The ``Time'' column gives the elapsed time at which the mean PSNR of the baseline reaches the feedforward PSNR of Gaussifier at that count. We interpolate it logarithmically between the budgets of Table~\ref{tab:budget_psnr}. Gaussifier takes at most 14\,ms, and the speedups use this value. The ``Reached'' column counts the images whose own curve reaches the PSNR of Gaussifier within 20\,s. The fitting runs use 50k Gaussians for Fast-2DGS on both datasets, on an RTX 5090. Our PSNR is measured as in Table~\ref{tab:matched}.}
\label{tab:matched_fitting}
\small
\setlength{\tabcolsep}{6pt}
\begin{tabular}{@{}llrcrrc@{}}
\toprule
Dataset & Baseline & \# Gaussians & Our PSNR & Time & Speedup & Reached \\
\midrule
\multirow{5}{*}{Kodak-24}
& AIR ps5    & 35,386 & 39.86 & 0.81\,s & 58$\times$  & 24/24 \\
& AIR ps6    & 25,468 & 37.84 & 0.94\,s & 67$\times$  & 24/24 \\
& AIR ps7    & 19,260 & 35.70 & 0.71\,s & 51$\times$  & 24/24 \\
& Instant-GI & 25,126 & 37.83 & 0.87\,s & 62$\times$  & 24/24 \\
& Fast-2DGS  & 50,000 & 41.56 & 4.26\,s & 304$\times$ & 22/24 \\
\midrule
\multirow{5}{*}{DIV2K}
& AIR ps5    & 36,294 & 34.90 & 1.21\,s & 86$\times$  & 100/100 \\
& AIR ps6    & 25,706 & 32.14 & 0.85\,s & 61$\times$  & 99/100 \\
& AIR ps7    & 19,260 & 29.74 & 0.36\,s & 26$\times$  & 98/100 \\
& Instant-GI & 28,825 & 33.11 & 0.60\,s & 43$\times$  & 100/100 \\
& Fast-2DGS  & 50,000 & 37.07 & 4.56\,s & 326$\times$ & 71/100 \\
\bottomrule
\end{tabular}
\end{table}

\subsection{Paired Statistics}
\label{app:paired_stats}
Table~\ref{tab:paired_stats} pairs the per-image PSNR of the methods in Tables~\ref{tab:main_table} and~\ref{tab:ablation_components}.
The PSNR of each method is first averaged over its seeds for each image.
The intervals are 95\% percentile bootstrap intervals over images with 10,000 resamples.
The fine-tuned baselines and the retrained rows of Table~\ref{tab:ablation_components} are not included.
\begin{table}[!htbp]
\centering
\caption{Paired PSNR differences in dB, with 95\% bootstrap intervals and the number of images on which the first method has the higher PSNR.}
\label{tab:paired_stats}
\small
\setlength{\tabcolsep}{4pt}
\begin{tabular}{@{}lrcrrcr@{}}
\toprule
& \multicolumn{3}{c}{Kodak-24} & \multicolumn{3}{c}{DIV2K} \\
\cmidrule(lr){2-4}\cmidrule(lr){5-7}
Comparison & Diff. & 95\% interval & Wins & Diff. & 95\% interval & Wins \\
\midrule
Gaussifier vs.\ AIR & +8.87 & [7.71, 10.07] & 24/24 & +11.04 & [10.44, 11.63] & 100/100 \\
Gaussifier vs.\ Instant-GI & +10.34 & [9.64, 11.06] & 24/24 & +11.20 & [10.73, 11.65] & 100/100 \\
Gaussifier vs.\ Fast-2DGS & +14.78 & [13.73, 15.92] & 24/24 & +14.88 & [14.33, 15.44] & 100/100 \\
Gaussifier-20k vs.\ AIR & +4.10 & [3.60, 4.60] & 24/24 & +2.58 & [2.22, 2.94] & 95/100 \\
Gaussifier-20k vs.\ Instant-GI & +5.57 & [4.64, 6.47] & 24/24 & +2.74 & [2.26, 3.22] & 90/100 \\
Gaussifier-20k vs.\ Fast-2DGS & +10.02 & [9.30, 10.73] & 24/24 & +6.42 & [6.01, 6.85] & 100/100 \\
\midrule
Full vs.\ uniform density & +5.87 & [4.52, 7.28] & 24/24 & +6.13 & [5.27, 7.01] & 100/100 \\
Full vs.\ gradient density & +1.28 & [1.03, 1.55] & 24/24 & +1.80 & [1.54, 2.09] & 98/100 \\
Full vs.\ RGB initialization & +8.93 & [7.68, 10.25] & 24/24 & +9.79 & [9.00, 10.58] & 98/100 \\
\bottomrule
\end{tabular}
\end{table}

\subsection{Perceptual Quality}
\label{app:lpips}
Table~\ref{tab:lpips} reports LPIPS~\citep{zhang2018lpips} with the AlexNet backbone for the methods of Table~\ref{tab:main_table}.
Gaussifier has the lowest LPIPS on both datasets.
With 20k Gaussians, it is on par with AIR on Kodak-24 and better on DIV2K.
\begin{table}[!htbp]
\centering
\caption{LPIPS with the AlexNet backbone for the feedforward methods of Table~\ref{tab:main_table}. Lower is better.}
\label{tab:lpips}
\small
\setlength{\tabcolsep}{8pt}
\begin{tabular}{@{}lcc@{}}
\toprule
Method & Kodak-24 & DIV2K \\
\midrule
Gaussifier & \textbf{0.0331} & \textbf{0.0143} \\
Gaussifier-20k & \underline{0.0676} & \underline{0.0751} \\
AIR & 0.0682 & 0.0987 \\
Instant-GI & 0.1594 & 0.1216 \\
Fast-2DGS & 0.3245 & 0.3119 \\
\bottomrule
\end{tabular}
\end{table}

\clearpage
\begin{figure}[p]
    \centering
    \includegraphics[width=\linewidth]{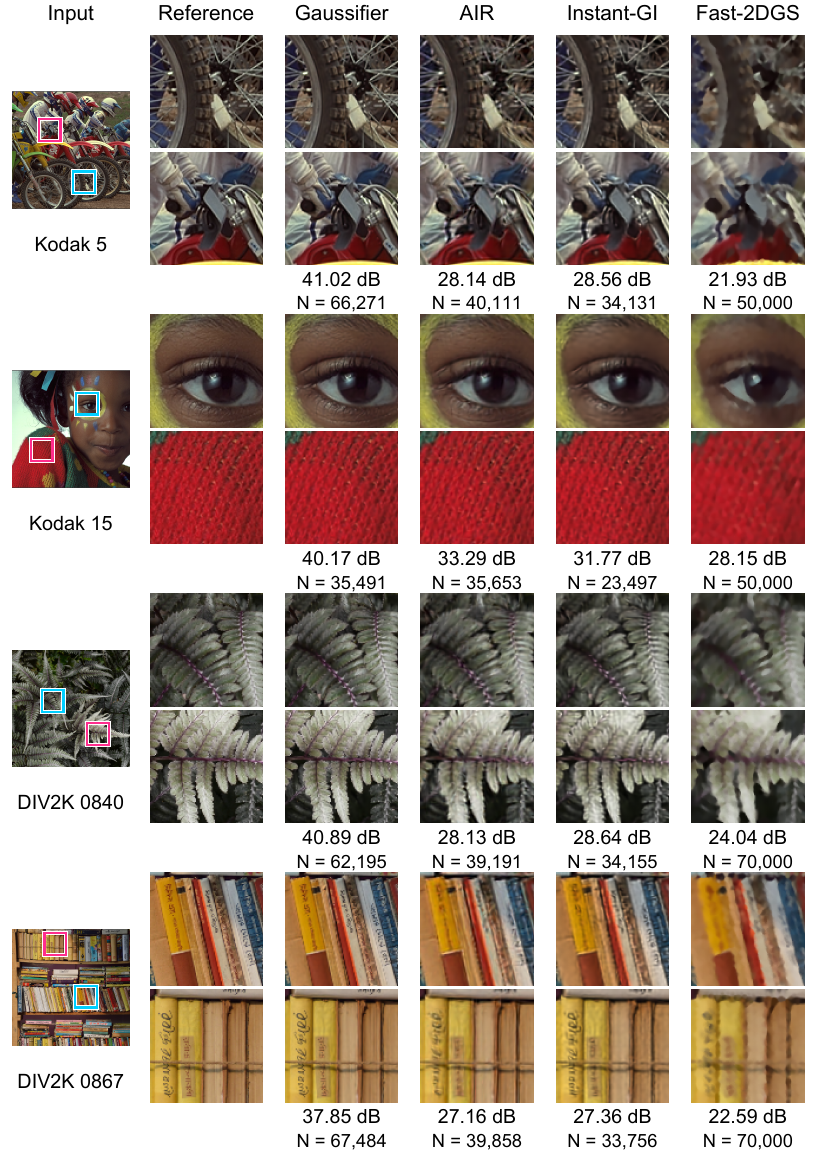}
    \caption{Additional feedforward reconstruction results. All models use the default settings of Table~\ref{tab:main_table}, with released baseline checkpoints and no per-image fitting. Each example shows the input and two corresponding $96\times96$ crops. Gaussifier uses its predicted count and six refinement updates. AIR uses ps5, and Fast-2DGS uses 50k/70k Gaussians on Kodak-24/DIV2K. PSNR and $N$ refer to full images. Fast-2DGS is displayed with seed~0 and its corresponding PSNR.}
    \label{fig:additional_default_reconstructions}
\end{figure}

\clearpage
\begin{figure}[p]
    \centering
    \includegraphics[width=\linewidth]{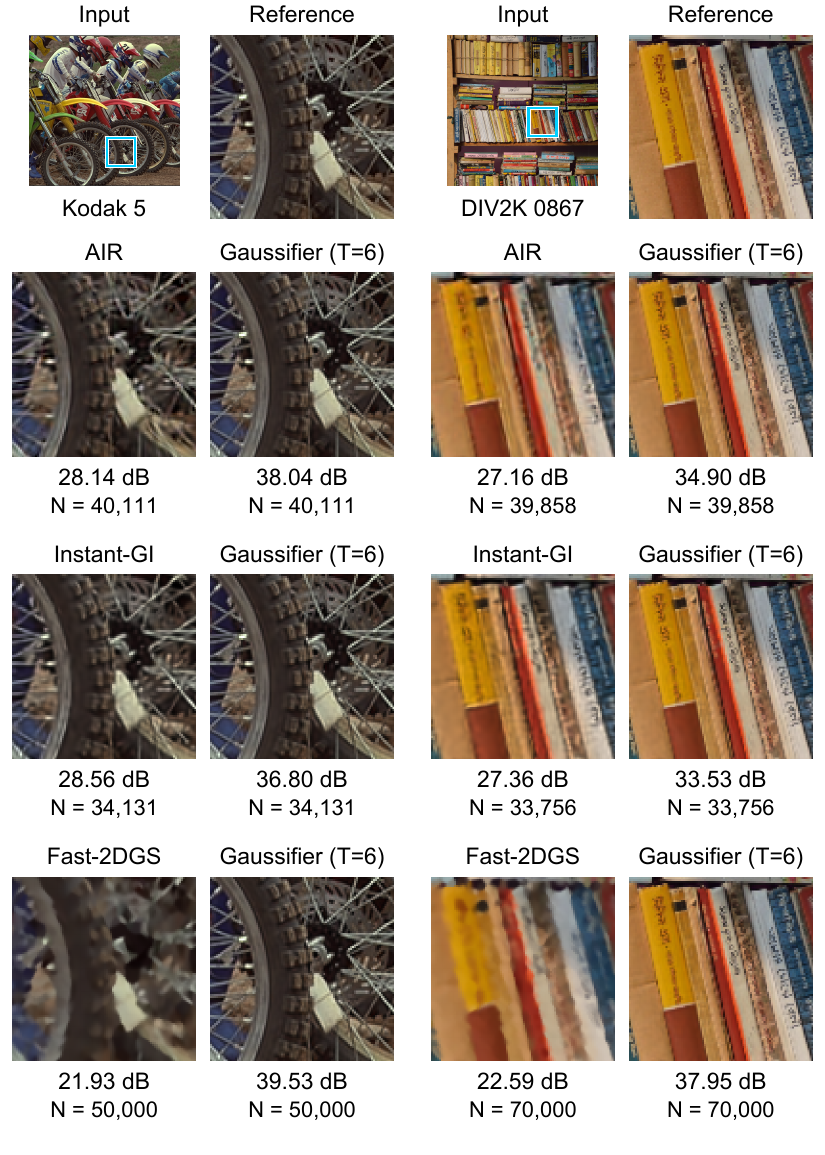}
    \caption{Reconstruction quality at matched Gaussian counts. The input and reference crop are shown once above each group of comparisons. For each image and baseline, Gaussifier uses the same Gaussian count and six refinement updates ($T=6$). Baselines use their released checkpoints, with AIR using ps5, and Gaussifier reconstructions are produced by its TensorRT/AOTInductor deployment. PSNR and $N$ refer to the full image. Fast-2DGS is displayed with seed~0 and its corresponding PSNR.}
    \label{fig:additional_matched_quality}
\end{figure}

\clearpage
\begin{figure}[p]
    \centering
    \includegraphics[width=\linewidth]{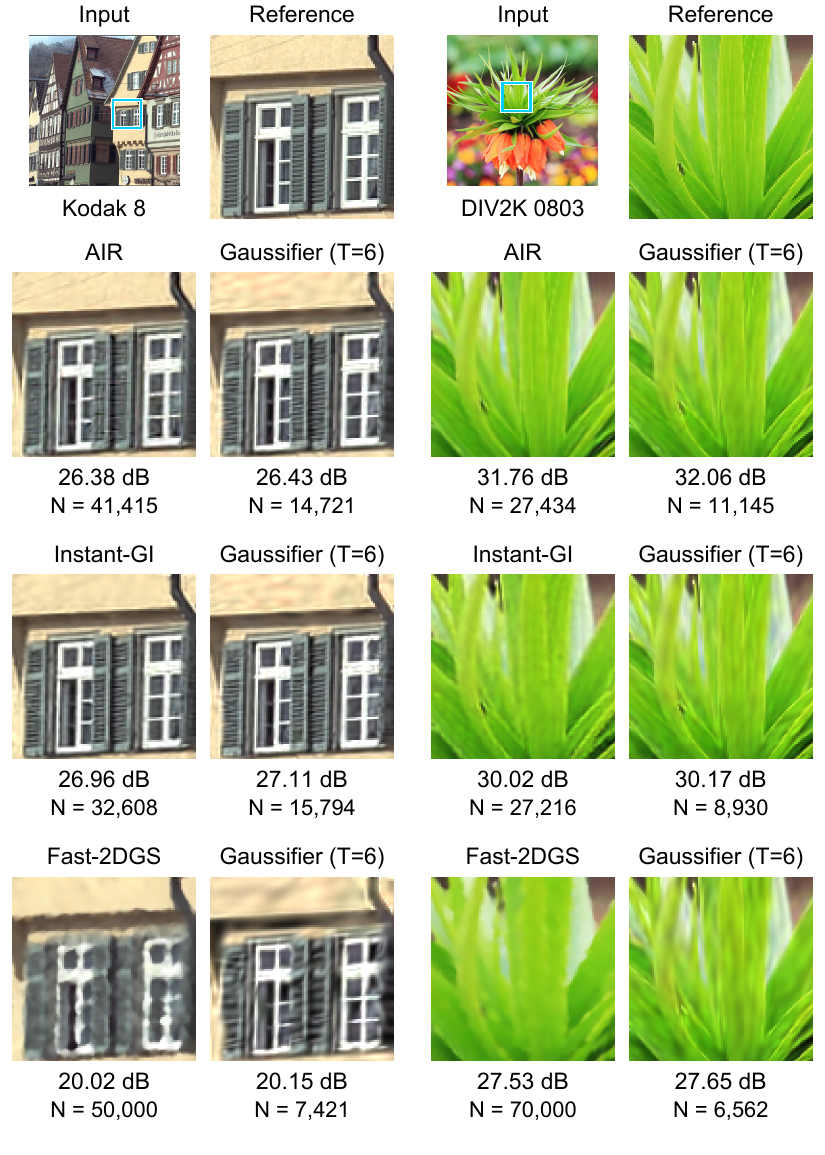}
    \caption{Gaussian counts at baseline reconstruction quality. The input and reference crop are shown once above each group of comparisons. For each image and baseline, we select the smallest evaluated Gaussian count whose full-image PSNR reaches the baseline target. Gaussifier uses six refinement updates ($T=6$) for every comparison, with reconstructions produced by its TensorRT/AOTInductor deployment. AIR uses the ps5 checkpoint. The Fast-2DGS target averages the PSNR of seeds 0--2, while its displayed reconstruction and PSNR use seed~0. PSNR and $N$ refer to full images.}
    \label{fig:additional_matched_count}
\end{figure}

\clearpage

\end{document}